\documentclass[final]{cvpr}

\usepackage{times}
\usepackage{epsfig}
\usepackage{graphicx}
\usepackage{amsmath}
\usepackage{amssymb}
\usepackage{color}
\usepackage{threeparttable}
\usepackage{booktabs}
\usepackage{wrapfig}
\usepackage{indentfirst}

\usepackage[pagebackref=true,breaklinks=true,colorlinks,bookmarks=false]{hyperref}

\def\cvprPaperID{8466} 
\def\confYear{CVPR 2021}

\begin{document}

\title{SSAN: Separable Self-Attention Network for Video Representation Learning}

\author{Xudong Guo$^{*1}$ \quad\quad\quad\quad Xun Guo$^2$ \quad\quad\quad\quad Yan Lu$^2$\\ 

$^1$Tsinghua University \quad\quad $^2$Microsoft Research Asia\\
{\tt\small  gxd20@mails.tsinghua.edu.cn  $\{$xunguo, yanlu$\}$@microsoft.com}\\

}
\maketitle
\pagestyle{empty}  
\thispagestyle{empty} 

\renewcommand{\thefootnote}{}
\footnotetext{$^*$The work was done when the author was with MSRA as an intern.}

\begin{abstract}
Self-attention has been successfully applied to video representation learning due to the effectiveness of modeling long range dependencies. Existing approaches build the dependencies merely by computing the pairwise correlations along spatial and temporal dimensions simultaneously. However, spatial correlations and temporal correlations represent different contextual information of scenes and temporal reasoning. Intuitively, learning spatial contextual information first will benefit temporal modeling. In this paper, we propose a separable self-attention (SSA) module, which models spatial and temporal correlations sequentially, so that spatial contexts can be efficiently used in temporal modeling. By adding SSA module into 2D CNN, we build a SSA network (SSAN) for video representation learning. On the task of video action recognition, our approach outperforms state-of-the-art methods on Something-Something and Kinetics-400 datasets. Our models often outperform counterparts with shallower network and fewer modalities. We further verify the semantic learning ability of our method in visual-language task of video retrieval, which showcases the homogeneity of video representations and text embeddings. On MSR-VTT and Youcook2 datasets, video representations learnt by SSA significantly improve the state-of-the-art performance. 
\end{abstract}

\section{Introduction}

\begin{figure}[t]
	\begin{center}
		\includegraphics[width=0.9\linewidth]{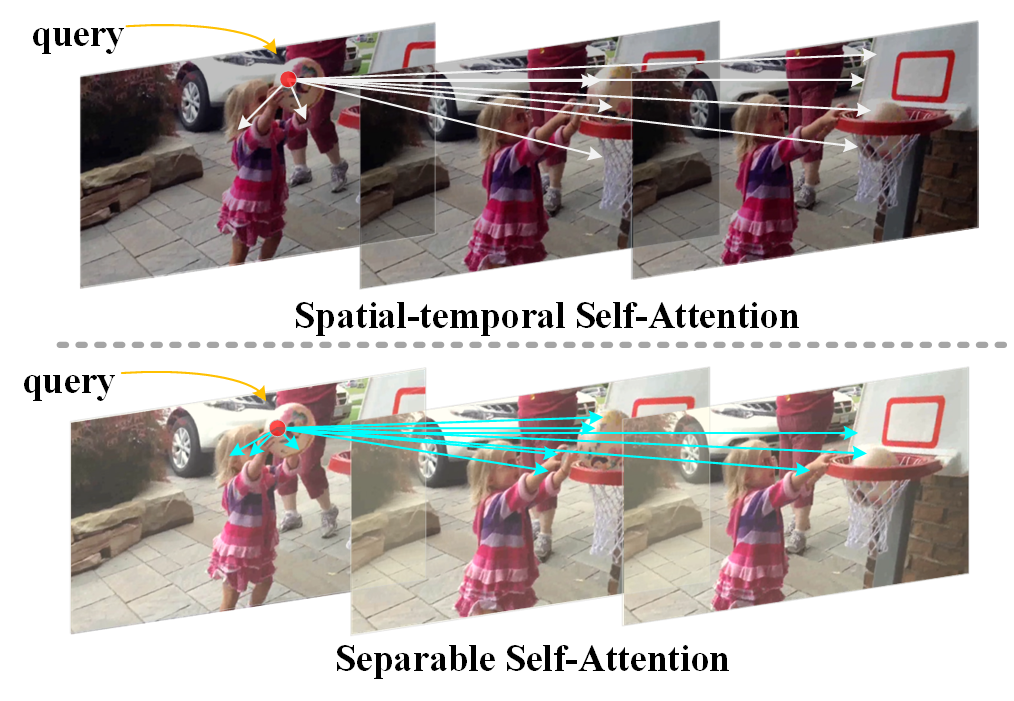}
	\end{center}
	\caption{Comparison between separable self-attention and spatial-temporal self-attention on Kinetics. The red points are query regions, and the arrows point to the relevant regions (Top-10 attention weights) with the query regions. Separable self-attention learns more action related regions  (hand and basket).}
	\label{fig:long}
	\label{fig:onecol}
\end{figure}

Video representation learning is crucial for tasks such as detection, segmentation and action recognition. Although 2D and 3D CNN based approaches have been extensively explored to capture the spatial-temporal correlations for these tasks, learning strong and generic video representations is still challenging. One possible reason is that videos contain not only rich semantic elements within individual frames, but also the temporal reasoning across time which links those elements to reveal the semantic-level information for actions and events. Effective modeling of long range dependencies among pixels is essential to capture such contextual information, which current CNN operations can hardly achieve. RNN based methods \cite{Alpher53} have been used for this purpose. However, they suffer from the high computational cost. More importantly, RNN cannot establish the direct pairwise relationship between positions regardless of their distance.

Self-attention mechanism has been recognized as an effective way to build long range dependencies. In natural language processing, self-attention based transformer \cite{Alpher08} has been successfully used to capture contextual information from sequential data, \textit{e.g.}, sentences. Recent efforts have also introduced self-attention to computer vision domain for visual tasks such as segmentation and classification \cite{Alpher44,Alpher45,Alpher16,Alpher48}. The work from Wang \textit{et al.} \cite{Alpher16} proposed a generic self-attention form, \textit{i.e.}, non-local mean, for video action recognition, which builds pairwise correlations for pixel locations from space and time simultaneously. However, the correlations from space and time represent different contextual information. The former often relates to scenes and objects, and the latter often relates to temporal reasoning for actions (short-term activities) and events (long-term activities). Human cognition always notices scenes and objects before their actions. Learning correlations along spatial and temporal dimensions together might capture irrelevant information, leading to the ambiguity for action understanding. This drawback becomes even worse for videos with complex activities. To efficiently capture the correlations in videos, decoupling the spatial and temporal dimensions is necessary. Meanwhile, short-term temporal dependencies should also be considered for capturing episodes of complex activities.\\
\indent In this paper, we fully investigate the relationship between spatial and temporal correlations in video, and propose a separable self-attention (SSA) module, which can efficiently capture spatial and temporal contexts for temporal modeling. In our design, spatial self-attention is first performed independently for input frames. The attention maps, which convey the spatial contextual information, are then aggregated along temporal dimension and sent to temporal attention module. In this way, the spatial contextual information will help better capture the temporal correlations for both short-term and long-term, so that the actions in videos can be fully understood.  \\
\indent We verify our approach on video action recognition task on Something-Something and Kinetics datasets. Something-Something (V1$\&$V2) contains fine-grained video action classes with high temporal reasoning, \textit{e.g.}, ``Moving something across a surface without it falling". By comparing with state-of-the-art 3D  \cite{Alpher01,Alpher12} and 2D \cite{Alpher05,Alpher11,Alpher19,Alpher20} based methods, our models show the superior performance. Moreover, our methods can outperform counterparts with shallower network structure, \textit{i.e.}, ResNet-50 vs. ResNet-101, and fewer modalities, \textit{i.e.}, RGB-only vs. RGB and optical flow. Since the goal of our design is to capture semantic information as well as possible, we further demonstrate the effectiveness of SSA in video-language task, \textit{i.e.}, video retrieval, which searches candidate video clips by text query. As text embeddings contain explicit semantic information, the homogeneity of video representations and text embeddings can better prove the efficiency of video learning method.

\begin{figure*}
	\begin{center}
		\includegraphics[width=1\linewidth]{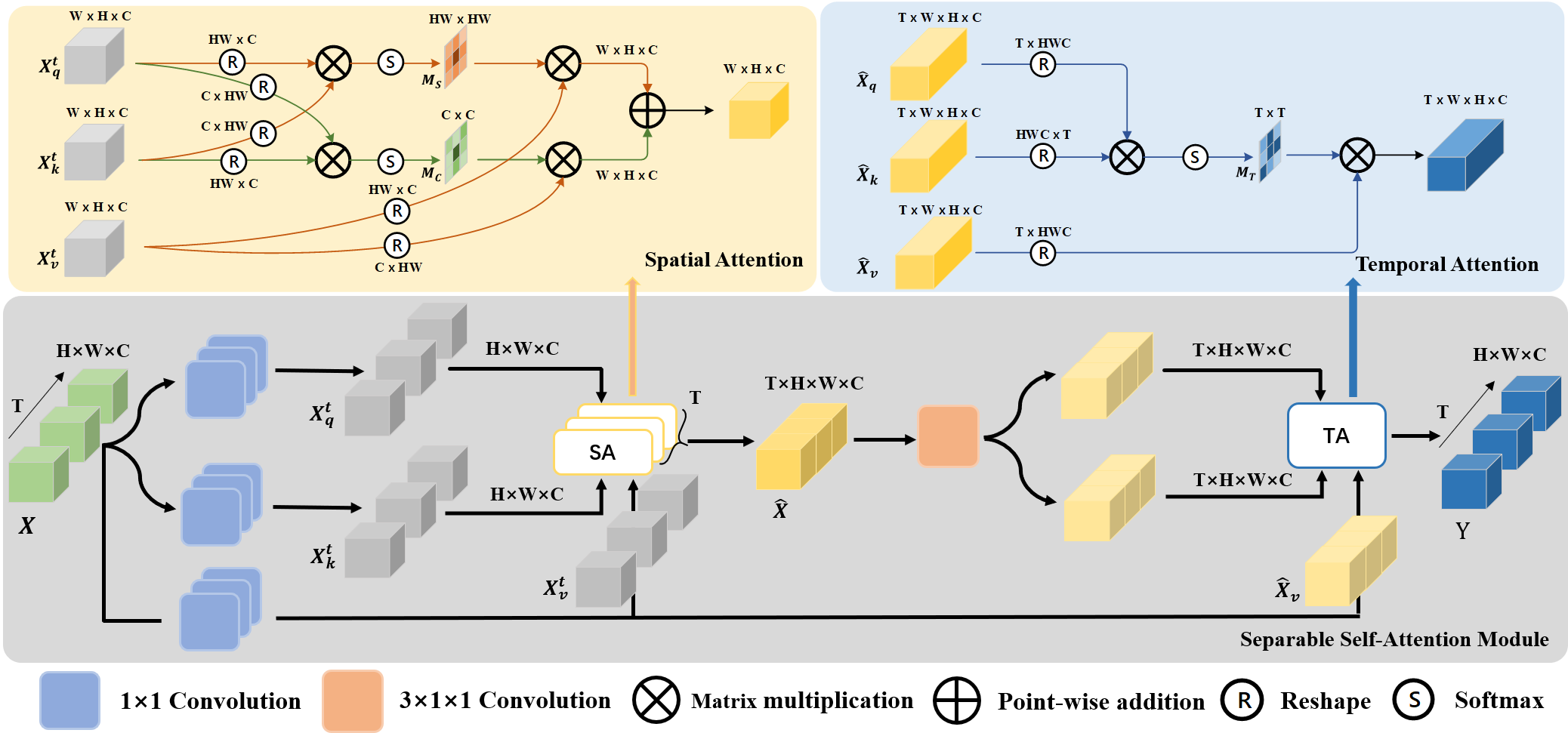}
	\end{center}
	\caption{Design of separable self-attention attention module. Spatial attention (SA) part is highlighted in yellow. Temporal attention (TA) part is highlighted in blue.}
	\label{fig:ssa}
\end{figure*}

\section{Related Work}

\subsection{Video Learning Networks}

With the success of CNN \cite{Alpher14,Alpher28,Alpher29,Alpher32} in image tasks, great efforts have been put on video tasks. CNN networks are also extended from image to video. There are mainly two branches for video learning architectures: 2D based methods \cite{Alpher37,Alpher11,Alpher23} and 3D based methods \cite{Alpher06,Alpher07,Alpher12,Alpher13}. I3D \cite{Alpher06} network proposed to inﬂate ImageNet pre-trained models from 2D to 3D by copying the weights. S3D \cite{Alpher12} proposed to decompose 3D convolutions to spatial and temporal convolutions. TSN \cite{Alpher11} proposed an eﬃcient 2D based video learning structure. And TSM \cite{Alpher05} proposed to capture temporal dependency by simply shifting the features between frames. SlowFast \cite{Alpher21} used two branches with diﬀerent temporal resolutions to capture temporal correlations. 3D based methods suffer from the overhead of parameters and complexity, while 2D based methods need careful temporal feature aggregation. To take the flexibility of 2D based methods for video frames, we employ 2D CNN as our baseline in this work. 
\subsection{Action Recognition}
Action recognition, also known as video classification \cite{Alpher10,Alpher14,Alpher17}, has been extensively explored in recent years. The accuracy of action recognition is highly related to video representation learning. Early works try to use 2D based methods to video. Later, 3D convolution networks are explored and achieve excellent performance. However, huge complexity makes these methods expensive to be used. Moreover, 3D based methods always take several consecutive frames as input, so that videos with complex actions can not be well handled. Recently, 2D convolution networks with temporal aggregation achieve significant improvements, which have flexible structures and inputs, as well as much lower computational and memory costs than 3D based methods. Representations learnt for action recognition are always used as initialization for other tasks.

\subsection{Self-Attention}
Self-attention mechanism has been successfully used in machine translation domain \cite{Alpher08}. Recently, there have been extensive eﬀorts to study its applications on computer vision tasks such as classification and segmentation \cite{Alpher44,Alpher45,Alpher16,Alpher22,Alpher36,Alpher46,Alpher43,Alpher48}. There are also efforts to use attention instead of convolutions for feature extraction \cite{Alpher47}. For video classification, non-local network \cite{Alpher16} proposed to use non-local mean, which is based on self-attention, to capture global dependency among pixel positions. Each pixel position attends to all other positions from both spatial and temporal dimensions. The full connections among pixels do take broader attention fields. However, this also introduces irrelevant information. Our work is closely related to non-local block, and try to investigate the relation of spatial attention and temporal attention. 

\subsection{Visual-Language Learning}
Recently, joint vision and language training \cite{Alpher27} is more and more popular for vision tasks. By adding language models, such as BERT \cite{Alpher33}, the semantic information can be efficiently learned from videos for multi-modal tasks, such as retrieval and captioning \cite{Alpher51,Alpher55}. VideoBERT \cite{Alpher25} proposed to use a visual-linguistic model to learn high-level features without any explicit supervision. Sun $et$ $al.$ \cite{Alpher26} then proposed to use contrastive bidirectional transformer (CBT) to perform self-supervised learning for video representations. UniViLM \cite{Alpher49} also proposed a joint video and language pre-training scheme by adding generation tasks in the pre-training process, which is more efficient than CBT in some downstream tasks. However, these methods only focus on the training of the transformer encoder and decoder, while take the video networks as feature extractors. Therefore, it is still important to improve video learning networks to efficiently learn real semantic information. 
\section{Separable Self-Attention Network}

In this section, we describe our proposed separable self-attention network in details.

\subsection{Self-Attention in Vision}
Let $X \in \mathbb{R}^{T \times H \times W \times C}$ be the input features with \textit{T} frames. \textit{H}, \textit{W} and \textit{C} denote spatial size, temporal size and channel number, respectively. A typical 3D self-attention/3D NL block \cite{Alpher16} maps \textit{X} into \textit{query}, \textit{key} and \textit{value} embeddings using three 1$\times$1$\times$1 convolutions, which are denoted as $X_q$, $X_k$ and $X_v$. The three embeddings are then reshaped to the sizes of $\textit{THW}\times\textit{C}$, $\textit{C}\times\textit{THW}$ and $\textit{THW}\times\textit{C}$, respectively. After that, the similarity matrix $M \in \mathbb{R}^{THW \times THW}$, which builds the full pairwise relationships for locations from spatial and temporal dimensions, is calculated using matrix multiplication as
\begin{equation}
\label{parwise}
M=X_q \times X_k.
\end{equation}
$M$ is then normalized with $softmax$ function and distributed to each location to generate the attention map $Y$ as
\begin{equation}
\label{value}
Y=softmax(M) \times X_v,
\end{equation}
where $Y \in \mathbb{R}^{THW \times C}$. Each element $m_{ij}$ in matrix $M$ measures the similarity between position $i$ and position $j$ in spatial and temporal dimensions. Attention $Y$ is then transformed by 1$\times$1$\times$1 convolution $W_z$ and added back to the original query feature $X$, like a residual connection, to generate the output feature $Z$:
\begin{equation}
\label{att}
Z=W_z(Y)+X.
\end{equation}
\subsection{Separable Self-Attention Module}
\indent The 3D based self-attention can successfully model long range dependencies from space and time simultaneously. However, such dependencies are first-order correlations which mainly capture the similarity between single pixels, but not semantic-level correlations. For example, if $i$ and $j$ are positions from different frames, lacking the prior spatial correlations between $i$ and other positions in the same frame with $i$, the calculated correlation between $i$ and $j$ may not describe the true temporal relationships of the scenes and objects they belong to. Furthermore, Equation \ref*{parwise} shows that the existing self-attention design, such as non-local block \cite{Alpher16}, considers more on position-wise correlations, but less on channel-wise correlations, which contain important classification information. This may lead to information loss for scenes and objects.  \\
\indent
Based on above analysis, we carefully design a separable self-attention module, which follows two principles. Firstly, the spatial and temporal attentions are performed sequentially, so that temporal correlations can fully consider the spatial contexts. Secondly, spatial attention maps exploit as much context information as possible.  

\indent The main structure of our proposed SSA is illustrated in Figure \ref{fig:ssa} highlighted in grey. The input feature $X$ is first mapped into spatial \textit{query}, \textit{key} and \textit{value} embeddings using 2D 1$\times$1 convolutions, denoted as $X_q^t$, $X_k^t$ and $X_v^t$, where $t\in[0,T]$ is time index. Then, the embeddings are used to generate $T$ spatial attention maps independently, which are then concatenated together as a 4D intermediate attention maps $\widehat{X}$ with the dimension of $T \times H \times W \times C$. After that, $\widehat{X}$ is transformed to temporal embeddings $\widehat{X}_q$ and $\widehat{X}_v$ using 3$\times$1$\times$1 convolution. Note that in our design, the spatial attention and temporal attention share the same \textit{value} embedding with different shapes, \textit{i.e.}, $X_v^t$ and $\widehat{X}_v$. The three temporal embeddings are then used to generate the temporal attention. We describe the details of spatial attention and temporal attention as follows.

\textbf{Spatial attention:} For spatial attention, we consider both position-wise attention and channel-wise attention. For this purpose, we design a two-branch structure, \textit{i.e.}, position branch and channel branch, to calculate them independently. The details are highlighted in yellow in Figure \ref{fig:ssa}. The two branches share the same embeddings $X_q^t$ and $X_k^t$. Each of the embeddings is reshaped to the size of $HW \times C$ and $C \times HW$. The position branch generates spatial similarity matrix $M_S \in \mathbb{R}^{N \times N}$, where $N=H \cdot W$. This is also the design of most existing 2D self-attention methods \cite{Alpher45}. The channel branch generates channel similarity matrix $M_C \in \mathbb{R}^{C \times C}$ to explore the dependencies along channel dimension. The two branches are calculated as
\begin{equation}
M_S=X_{q(S)}^t \times X_{k(S)}^t
\end{equation}
\begin{equation}
M_C=X_{q(C)}^t \times X_{k(C)}^t,
\end{equation}
where $S$ and $C$ denote position branch and channel branch, respectively. The spatial attention maps for time $t$ are then calculated as
\begin{equation}
\widehat{X}_t=(M_S \times X_{v(S)}^t) + (M_C \times X_{v(C)}^t).
\end{equation}
The intermediate attention maps $\widehat{X}=Cat[\widehat{X}_0, \widehat{X}_1...\widehat{X}_T]$ can be generated for the next stage, \textit{i.e.,} temporal attention.

\textbf{Temporal attention:} Temporal attention is performed using the spatial attention maps $\widehat{X}$ as input. Unlike the 3D self-attention that uses three 1$\times$1$\times$1 convolutions to generate the embeddings, we use one 3$\times$1$\times$1 convolution instead. This design allows temporal fusion on spatial attention maps and builds the short range correlations along temporal dimension, so that the short-term activities can also be attended to. The feature maps of $\widehat{X}$ are then reshaped to generate $\widehat{X}_q$ and $\widehat{X}_k$ for calculating similarity matrix $M_T \in \mathbb{R}^{T \times T}$ as
\begin{equation}
M_T=\widehat{X}_q \times \widehat{X}_k
\end{equation}
Thus, the final output attention map $Y$ is calculated as
\begin{equation}
Y=M_T \times \widehat{X}_v.
\end{equation}

\subsection{Network Architecture}

\begin{figure}[t]
	\begin{center}
		\includegraphics[width=1\linewidth]{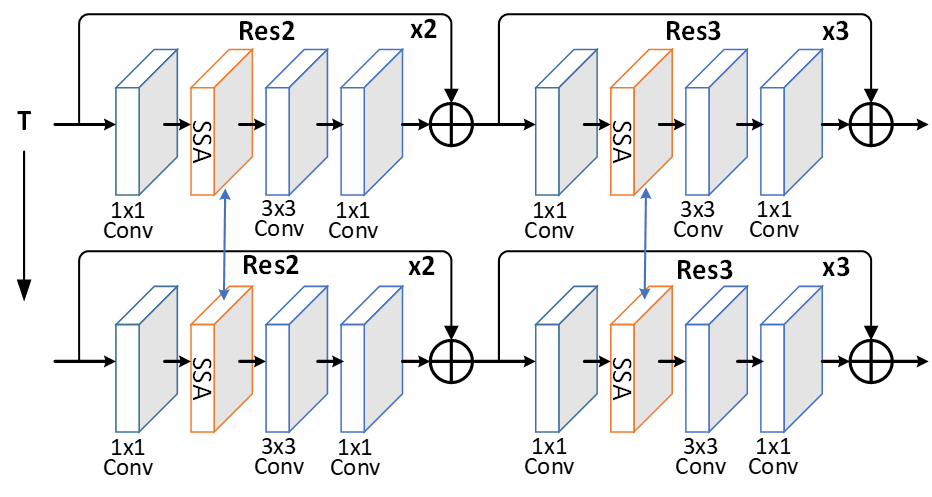}
	\end{center}
	\caption{Network architecture of SSA on ResNet-50. SSA module is inserted within residual block, $i.e.$, Res2 and Res3, right after the first 1$\times$1 convolution. There are totally 5 blocks containing SSA, 2 from Res2 and 3 from Res3. Temporal interactions of input frames only happen in SSA modules (blue arrow lines). The features are fused together at the end of the network.}
	\label{fig:network}
\end{figure}

The recent works on video learning show the excellent performance of 2D CNN based frameworks such as TSN \cite{Alpher11}, TRN \cite{Alpher20} and TSM \cite{Alpher05} . Comparing with the 3D CNN based methods, 2D based methods can better decouple the spatial and temporal modeling, thereby achieve superior performance on video clips with high dynamics. We choose TSN framework as our baseline, which uniformly splits a video clip into T snippets and selects only one frame per snippet. We insert SSA module into different layers to establish separable self-attention network (SSAN). Figure \ref{fig:network} shows an example of SSAN on ResNet-50 backbone. In this architecture, SSA module is inserted into the residual blocks to capture spatial and temporal contextual information in different stages. Particularly, we insert SSA module in Res2 and Res3 blocks right after the first 1$\times$1 convolution. Please note that there is temporal fusion only at the end of TSN. Thus, the temporal information exchange in middle layers comes only from SSA module, which better illustrates the temporal modeling ability of SSA. SSA is a flexible building module and can also be easily added in 3D based architecture.

\begin{table*}[t]
	\begin{center}
	\small
		\begin{threeparttable}
			\begin{tabular}{ccccccc}
				\toprule
				\textbf{Method} &  \textbf{Backbone} &    \textbf{Modality} & \textbf{Frames} & \textbf{Val Top-1} & \textbf{Val Top-5} & \textbf{Test Top-1}\\
				\midrule
				I3D \cite{Alpher06}    &   Inception & RGB & 64 & 45.8 & 76.5 & 27.2\\
				NL I3D + GCN \cite{Alpher06}    &   ResNet-50  & RGB  & 32+32 &  46.1 & 76.8 & 45.0\\
				S3D \cite{Alpher12}    &   Inception & RGB & 64 &   47.3 & 78.1 & -\\
				S3D-G \cite{Alpher12}    &   Inception & RGB & 64 & 48.2 & 78.7 & 42.0\\
				\midrule
				TSN \cite{Alpher11}    &   BNInception & RGB & 8 & 19.5 & - & -\\
				TRN \cite{Alpher20}    &   BNInception  & RGB  & 8 & 34.4 & - & 33.6 \\ 	
				\midrule
				bLVNet-TAM \cite{Alpher19}    &   bLResNet-50  & RGB  & 16$\times$2 & 48.4 & 78.8 & -\\
				bLVNet-TAM \cite{Alpher19}    &   bLResNet-101  & RGB  & 16$\times$2 & 49.6 & 79.8 & 48.9\\
				\midrule
				TSM \cite{Alpher05}    &   ResNet-50  & RGB  & 8 & 45.6 & 74.2 \\
				TSM \cite{Alpher05}    &   ResNet-50  & RGB  & 16 & 47.2 & 77.1 & 46.0\\
				TSM$_{En}$ \cite{Alpher05}    &   ResNet-50  & RGB  & 24 & 49.7 & 78.5 & - \\
				TSM 2-stream \cite{Alpher05}    &   ResNet-50  & RGB+flow & 16+16 & 52.6 & 81.9 & 50.7\\
				\midrule
				\midrule
				SSA(Ours)    &   ResNet-50  & RGB  & 8 & 49.5 & 79.5 & -\\  
				SSA(Ours)    &   ResNet-50  & RGB  & 16 & 51.7 & 81.3 & -\\ 
				SSA$_{En}$(Ours)    &   ResNet-50  & RGB  & 16+8 &  \textbf{55.1} & \textbf{84.9} & \textbf{54.0}\\
				\bottomrule
			\end{tabular}
		\end{threeparttable}
	\end{center}
	\caption{Comparisons with state-of-the-art methods on \textbf{Something-Something-V1} validation and test sets.}
	\label{tab:sthv1}
\end{table*}

\section{Experiments}
To demonstrate the effectiveness of our approach, we conduct comprehensive experiments on the standard vision task of video action recognition where the large-scale Something-Something dataset and Kinetics benchmark are used. Furthermore, to verify the efficiency of the video representations learnt by SSAN, we also conduct a vision-language task of text-based video retrieval which searches the corresponding video clips by text query. This task showcases the homogeneity between language and video representations, thereby the semantic-level information the video representations contain. 
\subsection{Dataset}
\textbf{Something-Something:} Something-Something dataset \cite{Alpher15} is a large scale benchmark dataset for video action recognition, containing 174 video categories of human-object interactions, \textit{e.g.}, “moving something across a surface without it falling”. Temporal reasoning is critical to infer the actions in this dataset. There are two versions, which contain 108k and 220k videos, respectively. Videos in Something-Something-V1 are split into 86K, 11K and 11K as training, validation and test sets. Something-Something-V2 is an updated version and contains 129K, 25K and 27K videos for training, validation and test. We conduct the experiments on both validation and test sets. It is noteworthy that labels for test sets are not publicly available. We submitted the inference results to the benchmark and report the scores published in the leaderboards. 

\textbf{Kinetics-400:} Kinetics-400 \cite{Alpher06} is a popular benchmark for action recognition collected from Youtube, which contains 400 action categories. There are totally 300K video samples, which are divided to 240K, 20K and 40K as training, validation and test sets, respectively. Videos in Kinetics-400 are relatively longer and more complex. Each video is trimmed to around 10-second clip. Kinetics is less sensitive to temporal relationships compared with Something-Something.

\subsection{Experimental Setups}

\textbf{Video action recognition:} We adopt the sparse sampling and data augmentation in TSN \cite{Alpher20} to train our model. Specifically, we first divide a video clip into T uniform segments and then select one random frame from each segment as the input. The input frames are resized as 256$\times$256 and randomly cropped to the size of 224$\times$224. We take single clip and the 224$\times$224 central crop for evaluation unless otherwise specified.

Our model is initialized with the weights pre-trained on ImageNet. For Something-Something dataset, we train our models for around 80 epochs. And for Kinetics-400 dataset, the models are trained for around 200 epochs. For all the models, the initial learning rate is 0.01 and decays by 0.1 when the validation loss reaches the plateau. We adopt a linear warmup strategy \cite{warmup} for the first epoch. Batch size is set as 64. We utilize the Nesterov momentum optimizer with a weight decay of 0.0005 and a momentum of 0.9. Dropout rate of 0.5 is also used to reduce over-fitting. 

\textbf{Text-based video retrieval:} This task aims at retrieving the most relevant video clip given an input text query. We adopt the method in \cite{howto}, which uses a gated recurrent unit (GRU) based text-video joint embedding network to measure the similarity between video representations and text embeddings. We adopt the training and inference strategies in \cite{howto} including text pre-processing, except that the video representations are extracted using our SSA with ResNet-50 instead of the original 3D ResNeXt-101 pre-trained on Kinetics dataset. We also pre-train our SSA with ResNet-50 on Kinetics dataset using video classification task. Then, we use SSA as a feature extractor to get video representations for retrieval. The GRU network can be pre-trained on HowTo100M \cite{howto} dataset, which is a video-language pre-training dataset containing one million narrated instructional web videos, then fine-tuned on MSR-VTT and Youcook2 datasets. The evaluation metric is ``Recall at K" (R@K), which is the percentage of video candidates retrieved in the top K video clips. We report this recall metrics on two popular  video-language datasets, $i.e.$, MSR-VTT and Youcook2. 
\begin{table}[t]
\small
	\begin{center}				
		\begin{tabular}{ccc}
			\toprule
			\textbf{Method} & \textbf{Backbone} & \textbf{Val Top1}\\
			\midrule
			3D NL & ResNet-50 & 60.6\\
			2D NL & ResNet-50 & 60.3\\
			1D NL & ResNet-50 & 60.1\\
			2D+1D NL & ResNet-50 & 61.0\\
			\midrule
			SSA & ResNet-50 & \textbf{62.3}\\
			\bottomrule
			
		\end{tabular}		
	\end{center}
	\caption{Comparisons with NL block on Something-Something-V2 validation set. All models use TSN based framework and ResNet-50 backbone initialized with the pre-trained weights on ImageNet. The input frame number is 8.}
	\label{tab:nl}
\end{table}
\subsection{Experiments on Action Recognition}
\textbf{Comparisons with non-local block:} Non-local block \cite{Alpher16} has been proved to significantly improve performance in video classification. We compare our approach with NL block to demonstrate the effectiveness of our separable design. In Table \ref{tab:nl}, we show the results of 3D NL (spatial-temporal), 2D NL (spatial) and 1D NL (temporal). We can see obvious performance improvement of SSA over 3D NL. This demonstrates that separating spatial and temporal attention is a right way to better model temporal reasoning. 

We also show the results of 2D+1D NL, which is a straightforward way to separate self-attention module along spatial and temporal dimensions. In specific, 1D NL is performed right after 2D NL. We insert NL block into the same locations with SSA for fair comparison, and observe superior performance of SSA. This interesting result shows that separable self-attention does need careful design. 

\textbf{Comparisons with state-of-the-art:} We compare our approach with previous state-of-the-art methods on Something-Something and Kinetics datasets. For Something-Something-V1 and V2, we report our results on both validation and test sets.

\begin{table}[t]
\small
	\begin{center}
		\begin{threeparttable}
			\begin{tabular}{cccc}
				\toprule
				\textbf{Method} & \textbf{Frames} &  \textbf{Val Top-1} & \textbf{Test Top-1} \\
				\midrule
				TSN  & 8 & 31.9 & -\\
				TRN$\dagger$ \cite{Alpher20} & 8 & 48.8 & 50.9 \\
				\midrule
				bLVNet-TAM \cite{Alpher19} & 16+16 & 61.7 & -\\
				bLVNet-TAM$\ddagger$ \cite{Alpher19}    & 16+16 & 61.9 &  -\\
				
				\midrule
				TSM \cite{Alpher05}   & 8 & 59.1 &  -\\
				TSM \cite{Alpher05}    & 16 & 63.4 &  64.3 \\
				TSM 2-stream \cite{Alpher05}    &   16+16 & 66.0 & 66.6\\
				
				\midrule
				\midrule
				SSA(Ours)    &   8 & 62.3 &  - \\  
				SSA(Ours)    &   16 & 66.0 &  - \\
				SSA$_{En}$(Ours)    & 16+8 & \textbf{67.4} & \textbf{68.2}\\
				\bottomrule
			\end{tabular}
		\end{threeparttable}
	\end{center}
	\caption{Comparisons with state-of-the-art 2D CNN based methods on \textbf{Something-Something-V2} validation and test sets.  $\dagger$ represents BNInception backbone, and $\ddagger$ represents ResNet-101.}
	\label{tab:sthv2}
\end{table}

\begin{table}[htbp]
\small
	\begin{center}
		\begin{threeparttable}
			\begin{tabular}{cccc}
				\toprule
				\textbf{Method}  & \textbf{Frames}  & \textbf{Val Top-1} & \textbf{Val Top-5} \\
				\midrule
				C3D$\dagger$ \cite{Alpher09}  &  -  & 65.6 & 85.7 \\
				I3D$\dagger$ \cite{Alpher06}    &  64 &  71.1 & 89.3 \\
				S3D$\dagger$ \cite{Alpher12}   &  64 & 72.2 & 90.6 \\
				\midrule
				TSN$\dagger$ \cite{Alpher11}    &  8 &  70.6 & 89.2 \\
				TSM \cite{Alpher05}    & 8 & 74.1 & 91.2 \\
				TSM \cite{Alpher05}    & 16 & 74.7 & - \\
				
				\midrule
				bLVNet-TAM \cite{Alpher19}     & 8+8 &  71.0 & 89.8 \\
				bLVNet-TAM \cite{Alpher19}   &  16+16 & 72.0 & 90.6 \\
				bLVNet-TAM \cite{Alpher19}    &    24+24 & 73.5 & 91.2 \\
				\midrule			
				$A^2$-Nets \cite{a2net} & 8 & 74.6 & 91.5\\
				NL I3D \cite{Alpher16} & 8 & 73.8 & 91.0\\
				NL I3D \cite{Alpher16} & 128 & 76.5 & 92.6 \\			
				\midrule
				\midrule
				SSA(Ours)   &   8 & 75.8 & 92.4 \\
				SSA(Ours)   &  16 & 76.4 & 92.7 \\
				SSA$_{En}$(Ours)    &  8+16 & \textbf{77.5} & \textbf{93.3} \\
				\bottomrule
			\end{tabular}
			\scriptsize
		\end{threeparttable}
	\end{center}
	\caption{Comparisons with state-of-the-art methods on \textbf{Kinetics-400}. The methods with $\dagger$ adopt Inception as backbone, while others adopt ResNet-50 as backbone.}
	\label{tab:k400}
\end{table}

The results of SSA and other state-of-the art methods on \textbf{Something-Something-V1} are summarized in Table \ref{tab:sthv1}. The first section shows the results of I3D and S3D, as well as their enhanced variants NL I3D + GCN and S3D-G, which enable full temporal fusion and attention mechanism. With much fewer input frames, our 16-frame model outperforms these methods by a large margin (5.9\%, 4.3\%, 5.6\% and 3.2\%, respectively). 
The second section shows the results of TSN and TRN. TSN has no temporal fusion operation, so that the performance is much lower. Our approach is built upon TSN framework. From the table, we can see a significant improvement of SSA (30.0\%) when using 8-frame input. Though TRN has a temporal fusion at the end of trunk network, the performance is still relatively low. This shows the fact that modeling temporal context across different layers is important. That is also why we choose to insert SSA module into the layers of the backbone, but not add it at the end as a head. Our RGB-only ensemble model ($SSA_{En}$), which is the ensemble of 8-frame input and 16-frame input, outperforms TSM 2-stream model with fewer input frames (24 $vs.$ 32). 
\begin{table}[t]	
\small
	\begin{center}	
		\begin{tabular}{ccccc}
			\toprule
			\textbf{2D+1D NL} & \textbf{SA} & \textbf{SVE} & \textbf{TA} & \textbf{Val Top1}\\
			\midrule
			\checkmark & & & & 61.0\\
			\checkmark & \checkmark & & & 61.5\\
			\checkmark & \checkmark & \checkmark & & 61.4\\
			\checkmark & \checkmark & \checkmark &  \checkmark & 62.3\\
			\bottomrule
			
		\end{tabular}		
	\end{center}
	\caption{Ablation study of individual performance for spatial attention (SA), temporal attention (TA) and shared value embedding (SVE) on Something-Something-V2. }
	\label{tab:sata}	
\end{table}

\begin{table}[t]	
\small
	\begin{center}	
		\begin{tabular}{ccccc}			
			\toprule
			\textbf{Method} & \textbf{Backbone} & \textbf{Res2} & \textbf{Res3} & \textbf{Val Top1}\\
			\midrule
			SSA & ResNet-50 & \checkmark & & 61.2\\
			SSA & ResNet-50 & & \checkmark & 61.8\\
			SSA & ResNet-50 & \checkmark &  \checkmark & 62.3\\
			\bottomrule			
		\end{tabular}		
	\end{center}
	\caption{Ablation study of different locations that SSA is inserted in. Note that we insert SSA into 5 residual blocks (2 from Res2 and 3 from Res3).}
	\label{tab:location}	
\end{table}

We also compare our approach on \textbf{Something-Something-V2} with state-of-the-art 2D CNN based methods in Table \ref{tab:sthv2}. Compared with Something-Something-V1, the videos in Something-Something-V2 are more sensitive to the number of input frames. It can be seen that SSA model with 16-frame input achieves comparable performance with TSM 2-stream model with 32-frame input. And compared with bLVNet-TAM, which has two branches of different spatial resolutions, our method achieves much better results.

In both Table \ref{tab:sthv1} and Table \ref{tab:sthv2}, SSA outperforms deeper network (ReNet-101) by a large margin (4.5\% and 5.5\% respectively), which show the strong video learning ability.

\begin{figure*}[t]
	\begin{center}
		\includegraphics[width=\linewidth]{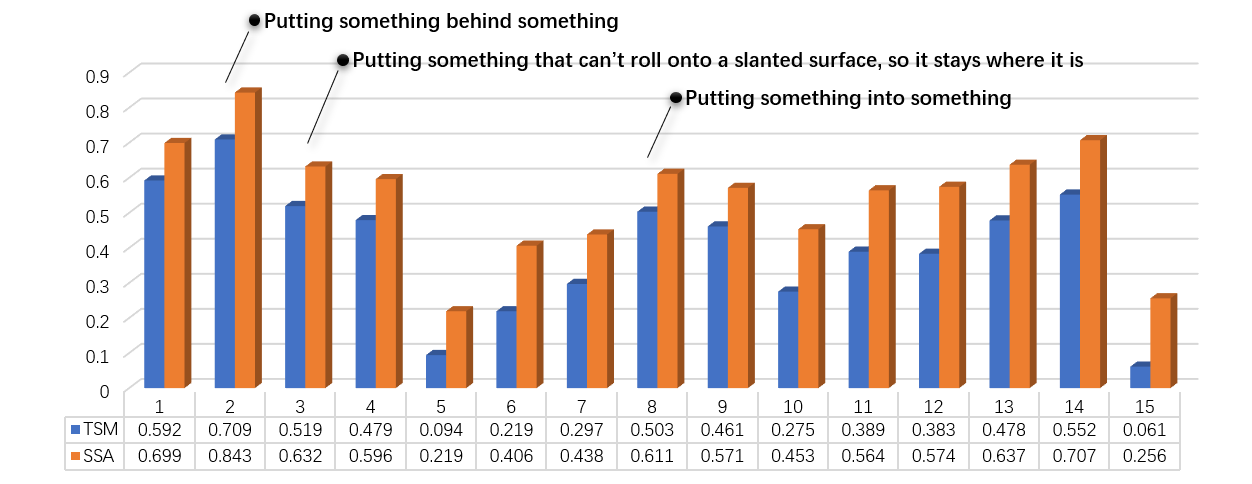}
	\end{center}
	\caption{The \textbf{top 15} categories which are improved by SSA (orange) compared to TSM (blue) On Something-Something-V2. There are some fine-grained categories that contain challenging actions such as ``\textbf{Putting something that can't roll onto a slanted surface, so it stays where it is}". Our model also works better on some categories with very small differences such as ``\textbf{Putting something behind something} and \textbf{Putting something into something}". }
	\vspace{-0.1cm}
	\label{fig:vis}
\end{figure*}

We also show per class results in Figure \ref{fig:vis} to figure out the efficiency of our method in individual classes. We list the results of TSM, which uses temporal shifting to exchange information among frames, to better understand the difference between temporal fusion and temporal attention. 

The performance comparisons on Kinetics-400 are summarized in Table \ref{tab:k400}, including two recent self-attention based methods, $i.e.$, $A^2$-Nets \cite{a2net} and NL block \cite{Alpher16}. Since Kinetics-400 dataset is less sensitive to the depth of network, we compare the results mainly on ResNet-50 backbone. From the table, we can see that our models outperform other methods when the numbers of input frames are the same. Our result with 16-frame input is comparable with that of NL I3D with 128-frame input, which demonstrates the strong temporal modeling ability of SSA. And our ensemble model ($SSA_{En}$) achieves 1\% higher accuracy than NL I3D with much fewer input frames (24 $vs.$ 128).

\textbf{Ablation study:} Although SSA has been proven to be efficient for video representation learning, we still like to fully investigate and understand this idea. Therefore, we conduct ablation experiments to showcase how each design affects the overall performance. 

As shown in Figure \ref{fig:ssa}, our design consists of two sequential parts, $i.e.$, spatial attention (SA) and temporal attention (TA). SA has a two-branch structure, which contains position branch and channel branch. The position branch is similar to 2D NL block. Therefore, SA is an enhanced version of 2D NL block by adding channel branch. In TA, we use 3$\times$1$\times$1 convolution instead of 1$\times$1$\times$1 convolution in NL block. We use 2D+1D NL as baseline and add our modifications one by one to demonstrate their contributions to the overall performance. Besides SA and TA, we also evaluate the performance of shared value embedding (SVE), which aims to reduce complexity by removing a 1$\times$1$\times$1 convolution. The ablation study is shown in Table \ref{tab:sata}. The results show that both SA and TA improve performance. When using them together, there is significant performance boost. Table \ref{tab:location} shows the results of different locations that SSA is inserted in. We can see that inserting SSA to either one of Res2 and Res3 can take significant improvement. According to our experiments, inserting SSA to the later blocks such as Res4/Res5 has lower performance than Res2/Res3 (around 0.5\%-0.7\%). This may be due to the reduction of spatial resolution.

\subsection{Experiments on Video Retrieval}

Video retrieval is a standard video-language task to find video candidates by text query. The normal process is to jointly train a cross encoder of video representations and text embeddings to learn their similarity. Text embeddings contain explicit semantic information. As a consequence, the more semantic information that video representations contain, the better the accuracies are. That is also major reason that we verify our method on this task. We employ two popular large-scale video-language datasets, $i.e.$, MSR-VTT and Youcook2. MSR-VTT contains videos in 20 domains such as music, sports and movies. For each video, 20 captioning sentences are annotated by human workers. There are totally 200K unique clip-caption pairs. We adopt the test strategy of JSFusion \cite{jsfusion} and use 1000K clip-caption pairs as test data. Youcook2 is a cooking video data set, including 14K video clips from 89 recipes. Since the video clips are much longer (5.26 minutes on average) and contain some uncorrelated scenes other than cooking instructions, Youcook2 is a challenging dataset.

\begin{table}[t]
\small
	\begin{center}
		\begin{tabular}{cccc}
			\toprule
			\textbf{Method} &
			\textbf{R@1$\uparrow$} & \textbf{R@5$\uparrow$} \\
			\midrule
			Random & 0.1 & 0.5 \\
			C+LSTM+SA \cite{Alpher51} & 4.2 & 12.9 \\
			VSE-LSTM \cite{vse} & 3.8 & 12.7 \\
			Kaufman \textit{et al.} \cite{kaufman} & 4.7 & 16.6 \\
			CT-SAN \cite{ctsan} & 4.4 & 16.6 \\
			JSFusion \cite{jsfusion} & 10.2 & 31.2 \\
			\midrule
			GRU+ResNeXt-101 \cite{howto} & 12.1 & 35.0 \\
			\midrule
			GRU+SSA & \textbf{24.4} & \textbf{49.3}\\
			\bottomrule
		\end{tabular}
	\end{center}
	\caption{Results of text-based video retrieval on \textbf{MSR-VTT} dataset. The only difference between our method and GRU+ResNeXt is the video representation.}
	\label{tab:msrvtt_r}
\end{table}

\begin{table}[t]
\small
	\begin{center}
		\begin{tabular}{cccc}
			\toprule
			\textbf{Method} &
			\textbf{R@1$\uparrow$} & \textbf{R@5$\uparrow$} \\
			\midrule
			Random & 0.03 & 0.15 \\
			HGLMM FV CCA \cite{cca} & 4.6 & 14.3 \\
			\midrule
			GRU+ResNeXt-101 \cite{howto} & 4.2 & 13.7 \\
			GRU+ResNeXt-101$\dagger$ \cite{howto} & 8.2 & 24.5 \\
			\midrule
			GRU+SSA & 5.5 & 15.9 \\
			GRU+SSA$\dagger$ & \textbf{10.9} & \textbf{28.4} \\
			
			\bottomrule
		\end{tabular}
	\end{center}
	\caption{Results of text-based video retrieval on \textbf{Youcook2} dataset. $\dagger$ represents pre-trained on a subset of HowTo100M dataset (around 1.2M videos).}
	\label{tab:youcook_r}
\end{table}

Table \ref{tab:msrvtt_r} summarizes the performance comparisons on MSR-VTT dataset. To demonstrate the affection of video features to the retrieval performance, no pre-training is applied on these methods. The first section shows previous state-of-the-art methods. From the results, we can see that text-to-video retrieval is really a challenging task. The second section shows the results from \cite{howto}, which our method outperforms by a large margin (12.3\%).

Table \ref{tab:youcook_r} summarizes the performance comparisons on Youcook2 dataset. The results show that pre-trained GRU using video representations learnt by SSA can be significantly improved (5.4\%), which is larger than 3D ResNeXt (4.0\%). This also shows that SSA is generic on different videos. 

It is noteworthy that MIL-NCE in~\cite{MILNCE} also reports the retrieval results on the two datesets. However, MIL-NCE is pre-trained on HowTo100M (around 1.2M videos), while SSA is pre-trained on Kinetics-400 (around 240K videos). Our experiments show that SSA outperforms MIL-NCE on MSR-VTT by 14.5\%, but is outperformed by MIL-NCE on Youcook2. Since they are not on the same basis, we don't include the comparison in above tables.


\section{Conclusion}
We proposed a separable self-attention network (SSAN) for video representation learning, which learns spatial and temporal correlations in a separable way. In specific, we investigate the relationship between spatial attention and temporal attention and design a sequential structure to model temporal reasoning with the priori of spatial contextual information. By adding the SSA module into 2D CNN backbone, we built a SSA network based on TSN framework. We conducted extensive experiments on large-scale Something-Something dataset and Kinetics-400 dataset to verify our approach. Our SSAN outperforms state-of-the-art methods on both datasets. Furthermore, we also verify the video representations learnt by our method on video-language task of video retrieval. On MSR-VTT and Youcook2 datasets, our method significantly improves state-of-the-art performance.

{\small
\bibliographystyle{ieee_fullname}
\bibliography{egbib}

\begin{thebibliography}{10}\itemsep=-1pt

\bibitem{Alpher47}
I. Bello, B. Zoph, A. Vaswani, J. Shlens, and Q.~V. Le.
\newblock Attention augmented convolutional networks.
\newblock In {\em International conference on computer vision (ICCV)}, 2019.

\bibitem{Alpher06}
J. Carreira and A. Zisserman.
\newblock Quo vadis, action recognition? a new model and the kinetics dataset.
\newblock In {\em Computer Vision and Pattern Recognition (CVPR)}, 2017.

\bibitem{a2net}
Y. Chen, Y. kalantidis, J. Li, S. Yan, and J. Feng.
\newblock $a^2$-nets: Double attention networks.
\newblock In {\em Neural Information Processing Systems (NIPS)}, 2018.

\bibitem{Alpher32}
F. Chollet.
\newblock Xception: Deep learning with depthwise separable convolutions.
\newblock In {\em Computer Vision and Pattern Recognition (CVPR)}, 2017.

\bibitem{Alpher33}
J. Devlin, M.-W. Chang, K. Lee, and K. Toutanova.
\newblock Bert: Pre-training of deep bidirectional transformers for language
  understanding.
\newblock {\em arXiv:1810.04805}, 2019.

\bibitem{Alpher19}
Q. Fan, C. Chen, H. Kuehne, M. Pistoia, and D. Cox.
\newblock More is less: Learning efficient video representations by big-little
  network and depthwise temporal aggregation.
\newblock In {\em Neural Information Processing Systems (NIPS)}, 2019.

\bibitem{Alpher21}
C. Feichtenhofer, H. Fan, J. Malik, and K. He.
\newblock Slowfast networks for video recognition.
\newblock In {\em International conference on computer vision (ICCV)}, 2019.

\bibitem{Alpher10}
C. Feichtenhofer, A. Pinz, and R. Wildes.
\newblock Spatiotemporal residual networks for video action recognition.
\newblock In {\em Neural Information Processing Systems (NIPS)}, 2016.

\bibitem{Alpher46}
J. Fu, J. Liu, H. Tian, Y. Li, Y. Bao, Z. Fang, and H. Lu.
\newblock Dual attention network for scene segmentation.
\newblock In {\em Computer Vision and Pattern Recognition (CVPR)}, 2019.

\bibitem{Alpher43}
R. Girdhar, J. Carreira, C. Doersch, and A. Zisserman.
\newblock Video action transformer network.
\newblock In {\em Computer Vision and Pattern Recognition (CVPR)}, 2019.

\bibitem{warmup}
P. Goyal, P. Dollar, R.~B. Girshick, P. Noordhuis, L. Wesolowski, A. Kyrola, A.
  Tulloch, Y. Jia, and K. He.
\newblock Accurate, large minibatch sgd: training imagenet in 1 hour.
\newblock In {\em CoRR}, 2017.

\bibitem{Alpher15}
R. Goyal, S.~E. Kahou, V. Michalski, J. Materzyńska, S. Westphal, H. Kim, V.
  Haenel, I. Fruend, P. Yianilos, M. Mueller-Freitag, F. Hoppe, C. Thurau, I.
  Bax, and R. Memisevic.
\newblock The something something video database for learning and evaluating
  visual common sense.
\newblock In {\em International conference on computer vision (ICCV)}, 2017.

\bibitem{Alpher36}
A. Harley, K. Derpanis, and I. Kokkinos.
\newblock Segmentation aware convolutional networks using local attention
  masks.
\newblock In {\em International conference on computer vision (ICCV)}, 2017.

\bibitem{Alpher48}
H. Hu, J. Gu, Z. Zhang, J. Dai, and Y. Wei.
\newblock Relation networks for object detection.
\newblock In {\em Computer Vision and Pattern Recognition (CVPR)}, 2018.

\bibitem{Alpher09}
S. Ji, W. Xu, M. Yang, and K. Yu.
\newblock 3d convolutional neural networks for human action recognition.
\newblock In {\em International Conference on Machine Learning (ICML)}, 2010.

\bibitem{Alpher37}
A. Karpathy, G. Toderici, S. Shetty, T. Leung, R. Sukthankar, and F. Li.
\newblock Large-scale video classiﬁcation with convolutional neural networks.
\newblock In {\em Computer Vision and Pattern Recognition (CVPR)}, 2014.

\bibitem{kaufman}
D. Kaufman, G. Levi, T. Hassner, and L. Wolf.
\newblock Temporal tessellation: A unified approach for video analysis.
\newblock In {\em International Conference on Computer Vision (ICCV)}, 2017.

\bibitem{vse}
R. Kiros, R. Salakhutdinov, and R. Zemel.
\newblock Unifying visual-semantic embeddings with multimodal neural language
  models.
\newblock {\em arXiv preprint arXiv:1411.2539}, 2014.

\bibitem{cca}
B. Klein, G. Lev, G. Sadeh, and L. Wolf.
\newblock Associating neural word embeddings with deep image representations
  using fisher vectors.
\newblock In {\em Computer Vision and Pattern Recognition (CVPR)}, 2015.

\bibitem{Alpher05}
J. Lin, C. Gan, and S. Han.
\newblock Tsm: Temporal shift module for efficient video understanding.
\newblock In {\em International conference on computer vision (ICCV)}, 2019.

\bibitem{MILNCE}
A. Miech, J.-B. Alayrac, L. Smaira, I. Laptev, J. Sivic, and A. Zisserman.
\newblock End-to-end learning of visual representations from uncurated
  instructional videos.
\newblock In {\em CVPR}, 2020.

\bibitem{howto}
A. Miech, D. Zhukov, J.-B. Alayrac, M. Tapaswi, I. Laptev, and J. Sivic.
\newblock How{T}o100{M}: {L}earning a {T}ext-{V}ideo {E}mbedding by {W}atching
  {H}undred {M}illion {N}arrated {V}ideo {C}lips.
\newblock In {\em International Conference on Computer Vision (ICCV)}, 2019.

\bibitem{Alpher53}
J.~Y. Ng, M. Hausknecht, S. Vijayanarasimhan, O. Vinyals, R. Monga, and G.
  Toderici.
\newblock Beyond short snippets: Deep networks for video classification.
\newblock In {\em Computer Vision and Pattern Recognition (CVPR)}, 2015.

\bibitem{Alpher23}
A. Piergiovanni, A. Angelova, and M.~S. Ryoo.
\newblock Tiny video networks.
\newblock {\em arXiv: 1910.06961}, 2019.

\bibitem{Alpher13}
Z. Qiu, T. Yao, and T. Mei.
\newblock Learning spatio-temporal representation with pseudo-3d residual
  networks.
\newblock In {\em International conference on computer vision (ICCV)}, 2017.

\bibitem{Alpher01}
Z. Qiu, T. Yao, C.-W. Ngo, X. Tian, and T. Mei.
\newblock Learning spatio-temporal representation with local and global
  diffusion.
\newblock In {\em Computer Vision and Pattern Recognition (CVPR)}, 2019.

\bibitem{Alpher49}
K. Schindler and L.~Van Gool.
\newblock Action snippets: how many frames does human action recognition
  require?
\newblock In {\em Computer Vision and Pattern Recognition (CVPR)}, 2008.

\bibitem{Alpher27}
W. Su, X. Zhu, Y. Cao, B. Li, L. Lu, F. Wei, and J. Dai.
\newblock Vl-bert: Pre-training of generic visual-linguistic representations.
\newblock In {\em ICLR}, 2020.

\bibitem{Alpher26}
C. Sun, F. Baradel, K. Murphy, and C. Schmid.
\newblock Learning video representations using contrastive bidirectional
  transformer.
\newblock {\em arXiv:1906.05743}, 2019.

\bibitem{Alpher25}
C. Sun, A. Myers, C. Vondrick, K. Murphy, and C. Schmid.
\newblock Videobert: A joint model for video and language representation
  learning.
\newblock In {\em International conference on computer vision (ICCV)}, 2019.

\bibitem{Alpher51}
A. Torabi, N. Tandon, and L. Sigal.
\newblock Learning language-visual embedding for movie understanding with
  natural-language.
\newblock {\em arXiv preprint arXiv:1609.08124}, 2016.

\bibitem{Alpher07}
D. Tran, L. Bourdev, R. Fergus, L. Torresani, and M. Paluri.
\newblock Learning spatiotemporal features with 3d convolutional networks.
\newblock In {\em International conference on computer vision (ICCV)}, 2015.

\bibitem{Alpher17}
D. Tran, H. Wang, L. Torresani, and M. Feiszi.
\newblock Video classification with channel-separated convolutional networks.
\newblock In {\em International conference on computer vision (ICCV)}, 2019.

\bibitem{Alpher14}
D. Tran, H. Wang, L. Torresani, J. Ray, Y. LeCun, and M. Paluri.
\newblock A closer look at spatiotemporal convolutions for action recognition.
\newblock In {\em Computer Vision and Pattern Recognition (CVPR)}, 2018.

\bibitem{Alpher08}
A. Vaswani, N. Shazeer, N. Parmar, J. Uszkoreit, L. Jones, A.~N. Gomez, L.
  Kaiser, and I. Polosukhin.
\newblock Attention is all you need.
\newblock In {\em Neural Information Processing Systems (NIPS)}, 2017.

\bibitem{Alpher11}
L. Wang, Y. Xiong, Z. Wang, Y. Qiao, D. Lin, X. Tang, and L.~Van Gool.
\newblock Temporal segment networks: Towards good practices for deep action
  recognition.
\newblock In {\em European Conference on Computer Vision (ECCV)}, 2016.

\bibitem{Alpher16}
X. Wang, R. Girshick, A. Gupta, and K. He.
\newblock Non-local neural networks.
\newblock In {\em Computer Vision and Pattern Recognition (CVPR)}, 2018.

\bibitem{Alpher28}
S. Xie, R. Girshick, P. Dollar, Z. Tu, , and K. He.
\newblock Aggregated residual transformations for deep neural networks.
\newblock In {\em Computer Vision and Pattern Recognition (CVPR)}, 2017.

\bibitem{Alpher12}
S. Xie, C. Sun, J. Huang, Z. Tu, and K. Murphy.
\newblock Rethinking spatiotemporal feature learning for video understanding.
\newblock In {\em European Conference on Computer Vision (ECCV)}, 2018.

\bibitem{Alpher44}
M. Yin, Z. Yao, Y. Cao, X. Li, Z. Zhang, S. Lin, and H. Hu.
\newblock Disentangled non-local neural networks.
\newblock In {\em European Conference on Computer Vision (ECCV)}, 2020.

\bibitem{jsfusion}
Y. Yu, J. Kim, and G. Kim.
\newblock A joint sequence fusion model for video question answering and
  retrieval.
\newblock In {\em European Conference on Computer Vision (ECCV)}, 2018.

\bibitem{ctsan}
Y. Yu, H. Ko, J. Choi, and G. Kim.
\newblock End-to-end concept word detection for video captioning, retrieval,
  and question answering.
\newblock In {\em IEEE Conference on Computer Vision and Pattern Recognition
  (CVPR)}, 2017.

\bibitem{Alpher29}
X. Zhang, X. Zhou, M. Lin, and Jian Sun.
\newblock Shufﬂenet: An extremely efﬁcient convolutional neural network for
  mobile devices.
\newblock In {\em Computer Vision and Pattern Recognition (CVPR)}, 2018.

\bibitem{Alpher20}
B. Zhou, A. Andonian, A. Oliva, and A. Torralba.
\newblock Temporal relational reasoning in videos.
\newblock In {\em European Conference on Computer Vision (ECCV)}, 2018.

\bibitem{Alpher55}
L. Zhou, Y. Zhou, J.~J. Corso, and R. Socher.
\newblock End-to-end dense video captioning with masked transformer.
\newblock In {\em Computer Vision and Pattern Recognition (CVPR)}, 2018.

\bibitem{Alpher22}
C. Zhu, X. Tan, F. Zhou, X. Liu, K. Yue, E. Ding, and Y. Ma.
\newblock Fine-grained video categorization with redundancy reduction
  attention.
\newblock In {\em European Conference on Computer Vision (ECCV)}, 2018.

\bibitem{Alpher45}
Z. Zhu, M. Xu, S. Bai, T. Huang, and X. Bai.
\newblock Asymmetric non-local neural networks for semantic segmentation.
\newblock In {\em International conference on computer vision (ICCV)}, 2019.

\end{thebibliography}
}

\end{document}


\title{SSAN: Separable Self-Attention Network for Video Representation Learning\\(Supplementary Material)}

\author{First Author\\
Institution1\\
Institution1 address\\
{\tt\small firstauthor@i1.org}
\and
Second Author\\
Institution2\\
First line of institution2 address\\
{\tt\small secondauthor@i2.org}
}

\maketitle

\section{Experiments on MobileNet-V2 Backbone}

\begin{table*}[!htbp]
	\begin{center}
		\begin{threeparttable}
			\begin{tabular}{cccccccc}
				\toprule
				Model   &  Backbone &    Pretrain & Frames  & Param (10$^{6}$) & FLOPs (10$^{9}$) &  Val Top-1 & Val Top-5 \\
				\midrule
				TSN (our impl.)   &   MobileNet-V2 & ImageNet & 8 & 2.4 & 2.5&16.5 & 41.6 \\
				TSM (our impl.)    &   MobileNet-V2  & ImageNet  & 8 & 2.4&2.5&41.5 & 71.1 \\
				\midrule
				\midrule
				SSA (Ours)    &   MobileNet-V2  & ImageNet  & 8 & 2.5 & 2.5 & \textbf{46.5} & \textbf{75.8} \\  
				\bottomrule
			\end{tabular}
		\end{threeparttable}
	\end{center}
\label{v1}
	\caption{Results on the Something-Someting-V1 validation set when using MobileNet-V2 as backbone.(center crop and 1 clip)}
\end{table*}

\begin{table*}[!htbp]
	\begin{center}
		\begin{threeparttable}
			\begin{tabular}{cccccccc}
				\toprule
				Model   &  Backbone &    Pretrain & Frames  & Param (10$^{6}$) & FLOPs (10$^{9}$) &  Val Top-1 & Val Top-5 \\
				\midrule
				TSN (our impl.)    &   MobileNet-V2 & ImageNet & 8 & 2.4 &2.5& 29.6&  60.1\\
				TSM (our impl.)    &   MobileNet-V2  & ImageNet  & 8 & 2.4& 2.5& 54.1& 81.7 \\
				\midrule
				\midrule
				SSA (Ours)    &   MobileNet-V2  & ImageNet  & 8 & 2.5 & 2.5 & \textbf{57.7} & \textbf{84.2} \\ 
				\bottomrule
			\end{tabular}
		\end{threeparttable}
	\end{center}
\label{v12}
	\caption{Results on the Something-Someting-V2 validation set when using MobileNet-V2 as backbone.(center crop and 1 clip)}
\end{table*}

In order to verify the consistent efficiency of our method, besides ResNet-50 \cite{Alpher08}, we also implemented our SSA module on MobileNet-V2 \cite{Alpher02}. MobileNet-V2 is much lighter than ResNet, which is more mobile devices friendly and could be easily applied to edge deployments. Inserted into MobileNet-V2, our SSA can still achieve significant improvements. Note that because there are no reported results of TSN \cite{Alpher07} and TSM \cite{Alpher03} MobileNet-V2 models on Something-Something datasets, we trained the models by ourselves for comparison. As shown in Table \ref{v1} and Table \ref{v2}, on both of the Something-Someting-V1\&V2 datasets, SSA outperforms TSM with almost no extra memory space and computation costs.

\section{Visualization}

We also provide the visualization results of our approach by comparing with NL. From Figure \ref{fig:ssa}, we can see that NL attends more to the objects, e.g., the plane and the hands, but less to their interactions. On the contrary, our method attend to both the objects and the interactions along temporal dimension.

\begin{figure*}
	\begin{center}
		\includegraphics[width=\linewidth]{vis.png}
	\end{center}
	\caption{Visualization of class activated maps (CAM) of SSA and NL. we choose 4 frames out of 8-frame inputs for this illustration. GT means the ground truth of the classes that the video clips belong to. In the middle row and the bottom row, heat maps of NL and SSA are listed respectively. NL gives wrong predictions for both of the two classes.}
	\label{fig:ssa}
\end{figure*}

{\small
\bibliographystyle{ieee_fullname}
\bibliography{smbib}
}